\documentclass[12pt]{article}
\usepackage{geometry}                
\usepackage{graphicx}
\usepackage{amssymb}
\usepackage{epstopdf}
\usepackage[tight]{subfigure}
\begin{document}

\title{Artificial life: sustainable self-replicating systems}
\author{Carlos Gershenson$^{1}$ \& Jitka \v{C}ejkov\'a$^{2}$\\
$^{1}$ Universidad Nacional Aut\'onoma de M\'exico. cgg@unam.mx\\
$^{2}$Vysok\'a \v{s}kola chemicko technologick\'a v Praze. Jitka.Cejkova@vscht.cz}
\date{}                                           

\maketitle
\section{General overview}

Nature has found one method of organizing living matter, but maybe other options exist --- not yet discovered --- on how to create life. To study the life ``as it could be'' is the objective of an interdisciplinary field called Artificial Life (commonly abbreviated as ALife)~\cite{langton1997artificial,Adami:1998,Aguilar2014The-Past-Presen}. The word “artificial” refers to the fact that humans are involved in the creation process. The artificial life forms might be completely unlike natural forms of life, with different chemical compositions, and even  computer programs exhibiting life-like behaviours.

\begin{figure}[ht]
    \centering
    \includegraphics[width=\textwidth]{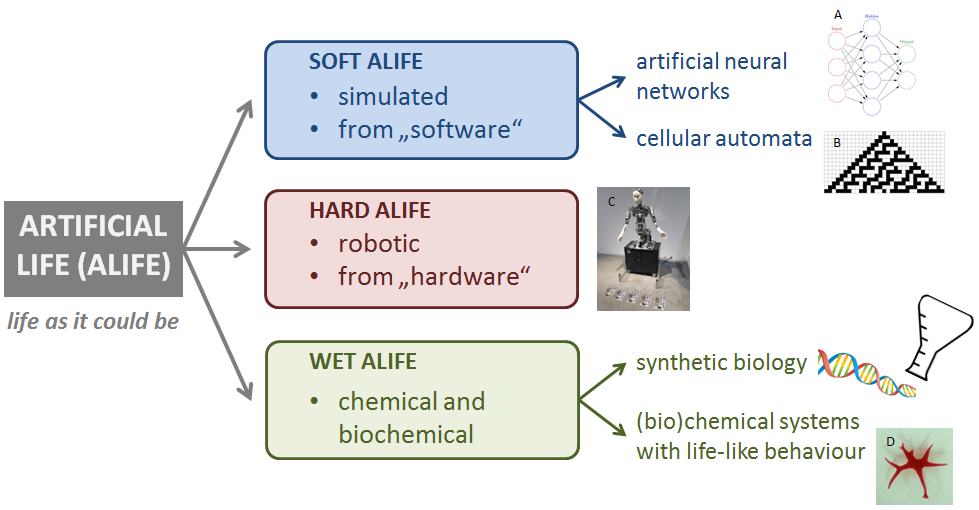}
    \caption{\label{fig:alife}Artificial life research. (A – Artificial neural networks. B – Cellular automaton, rule 30. C – Robot Alter 2. D – Shape changing decanol droplet)}
\end{figure}

ALife was established at the first “Interdisciplinary Workshop on the Synthesis and Simulation of Living Systems” in Los Alamos in 1987 by Christopher G. Langton~\cite{langton1987}. ALife is a radically interdisciplinary field that contains biologists, computer scientists, physicists, physicians, chemists, engineers, roboticists, philosophers, artists, and representatives from many other disciplines. There are several approaches to defining ALife research. One can discriminate between soft, hard and wet ALife (Figure~\ref{fig:alife}). \textbf{“Soft”} ALife is aiming to create simulations or other purely digital constructions exhibiting life-like behaviour. \textbf{“Hard”} ALife is related to robotics and implements life-like systems in hardware made mainly from silicon, steel and plastic. \textbf{“Wet”} ALife uses all kinds of chemicals to synthesize life-like systems in the laboratory. 

Bedau \emph{et al.}~\cite{BedauEtAl2000} proposed 14 open problems in ALife in the year 2000, but none of them have been solved yet. Aguilar \emph{et al.}~\cite{Aguilar2014The-Past-Presen} summarized the ALife research challenges and divided them into thirteen themes: origins of life, autonomy, self-organization, adaptation (evolution, development, and learning), ecology, artificial societies, behaviour, computational biology, artificial chemistries, information, living technology, art, and philosophy. 


\section{Soft ALife}

Mathematical and computational models of living systems are naturally more abstract than robotic or chemical systems. Also, they are easier to build, so actually most ALife models are ``soft''. Still, we can discriminate ``abstract'' and ``grounded'' soft ALife models. This distinction is more gradual than categorical. More abstract models focus less on physics or biology, and more on information and organization, while more physical models consider to a greater degree the actual components of living systems.

\begin{figure}
     \centering
     \subfigure[]{
          \label{fig:rbnA}
          \includegraphics[width=.44\textwidth]{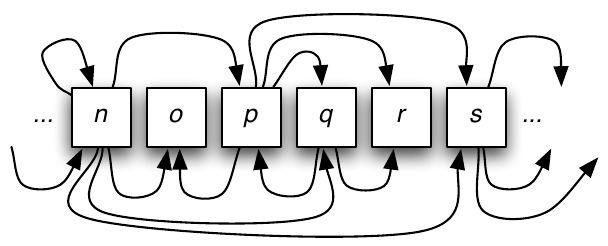}
	}
     \subfigure[]{
          \label{fig:rbnB}
          \includegraphics[width=.44\textwidth]{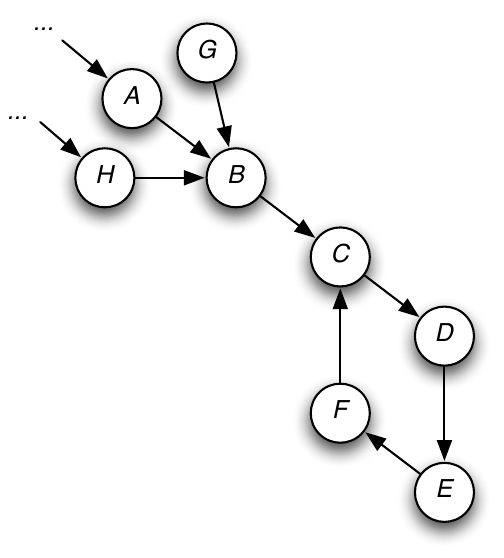}
	}\\
     \subfigure[]{
          \label{fig:rbnC}
          \includegraphics[width=.69\textwidth]{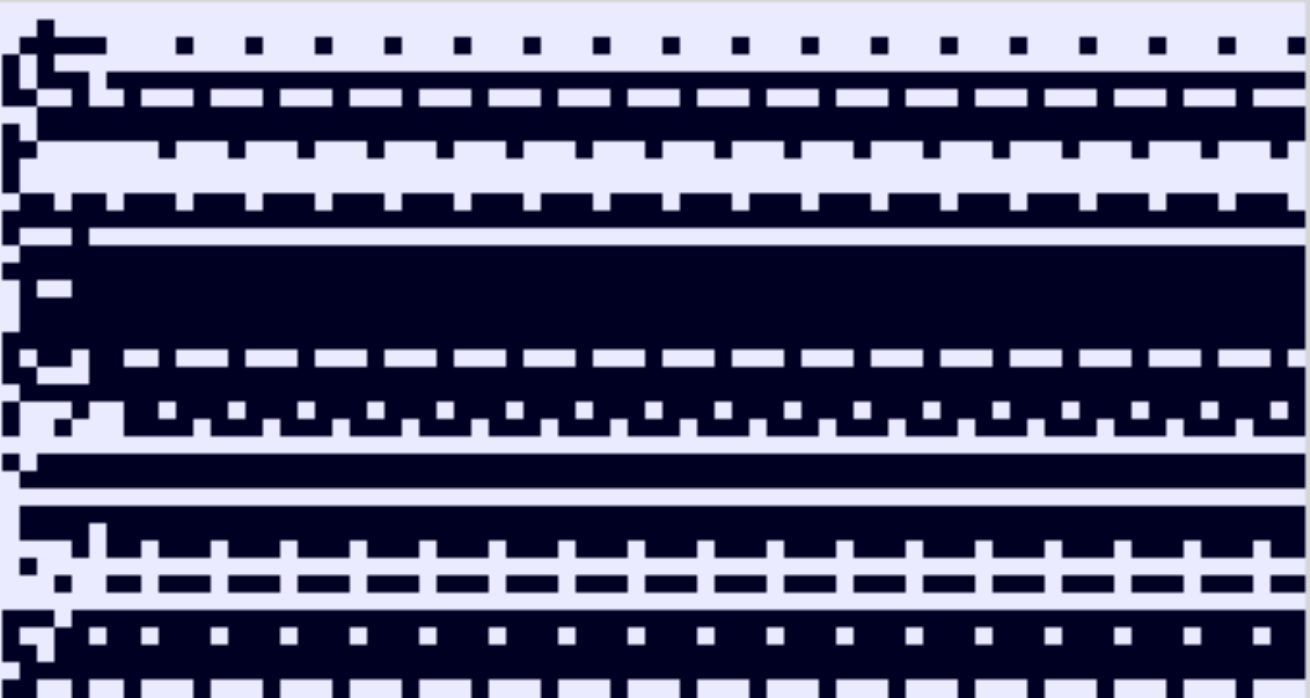}
	}

     \caption{An example of a random Boolean network~\cite{Gershenson:2010}. (A) A structural network is formed by $N$ Boolean nodes (can take values of zero or one) are connected to $K$ inputs randomly. The future state ($t+1$) of each node is determined by the current state ($t$) of its inputs following lookup tables, that are also generated randomly (and then remain fixed). (B) The structural network defines a state transition network with $2^N$ nodes. Each state has precisely one successor, but it can have several or no predecessors. Thus, it is a dissipative system. Eventually, a visited state is repeated, and thus the network has reached an attractor. (C) Example dynamics of a RBN with $N=40$ and $K=2$, time flowing to the right. A random initial state converges into an attractor of period 4. A single RBN can have several attractors of different periods.}
     \label{fig:rbn}
\end{figure}

\begin{figure}[ht]
    \centering
    \includegraphics[width=\textwidth]{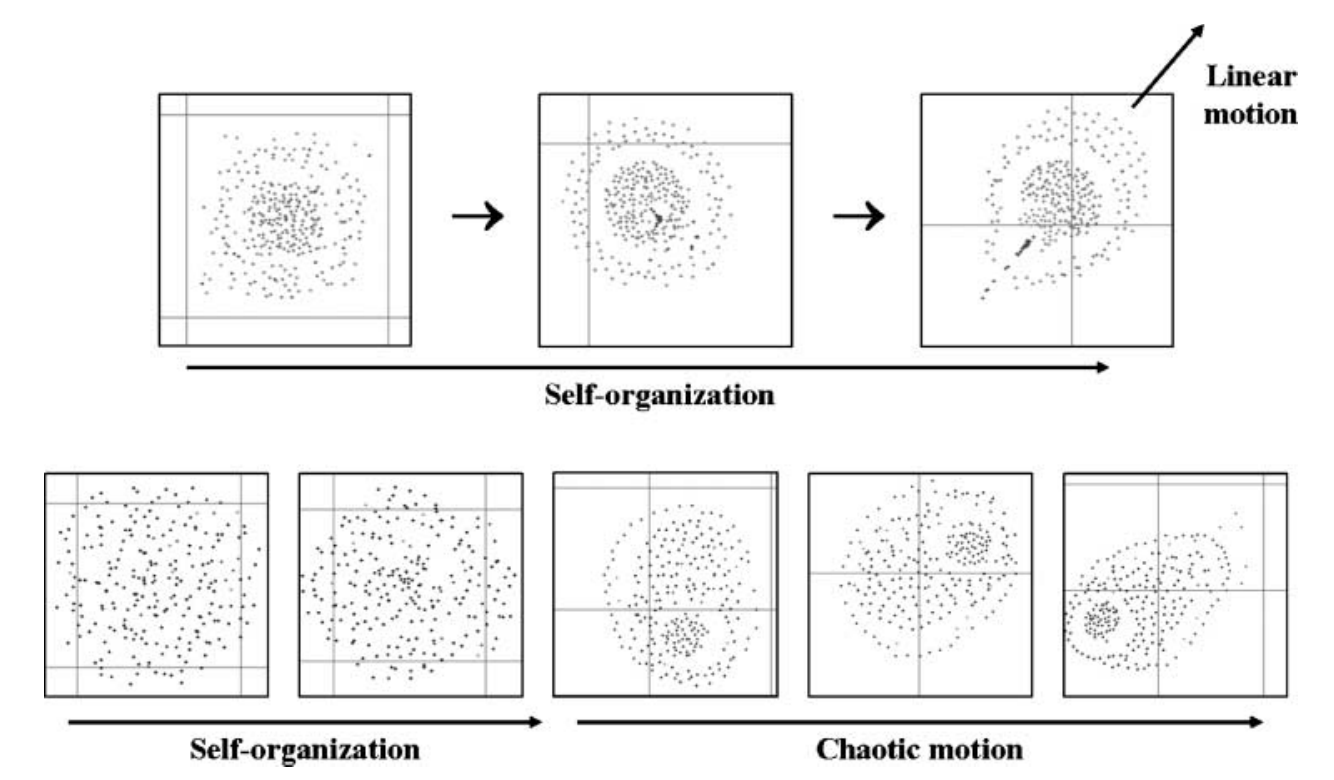}
    \caption{\label{fig:swarmChem}Examples of complex structures made of several different chemical species, designed using Sayama's interactive simulator~\cite{Sayama2008Swarm-Chemistry}. The swarms self-organize from initially random states to a shape that looks like a horseshoe crab (top), or a biological cell-like structure that shows active chaotic movement after self-organization (bottom).}
\end{figure}

Some of the most abstract models include cellular automata~\cite{vonNeumann1966,Shelling1971,BerlekampEtAl82,Wolfram1983,WuenscheLesser1992,Mitchell:1993,EpsteinAxtell1996}, random Boolean networks~\cite{Kauffman1969,AldanaEtAl2003,Gershenson2004c,Gershenson:2010} (Figure~\ref{fig:rbn}), and boids \cite{reynolds87flocks}, where there are basically space, time, and simple dynamics that may lead to complex behavior. Other abstract models have focussed on studying self-replication~\cite{Sipper98} or evolution \cite{Ray1994,AdamiBrown1994,Lenski1999Genome-complexi,Lenski2003The-evolutionar,Adami25042000}. Abstractly exploring the theoretical space of possibilities for living systems (necessary and sufficient conditions) has also been made at the ``chemical'' level, with artificial chemistries~\cite{fontana1991algorithmic,Dittrich2001Artificial-Chem} and swarm chemistry~\cite{Sayama2008Swarm-Chemistry} (Figure~\ref{fig:swarmChem}. Another more ``grounded'' strand of soft ALife involves the simulation of environments with realistic physics, either to evolve ``creatures'' \cite{Sims1994} or controllers for physical robots \cite{Jakobi1997,Lipson2000,Bongard:2006}.

\section{Hard ALife}

\begin{figure}[ht]
    \centering
    \includegraphics[width=0.6\textwidth]{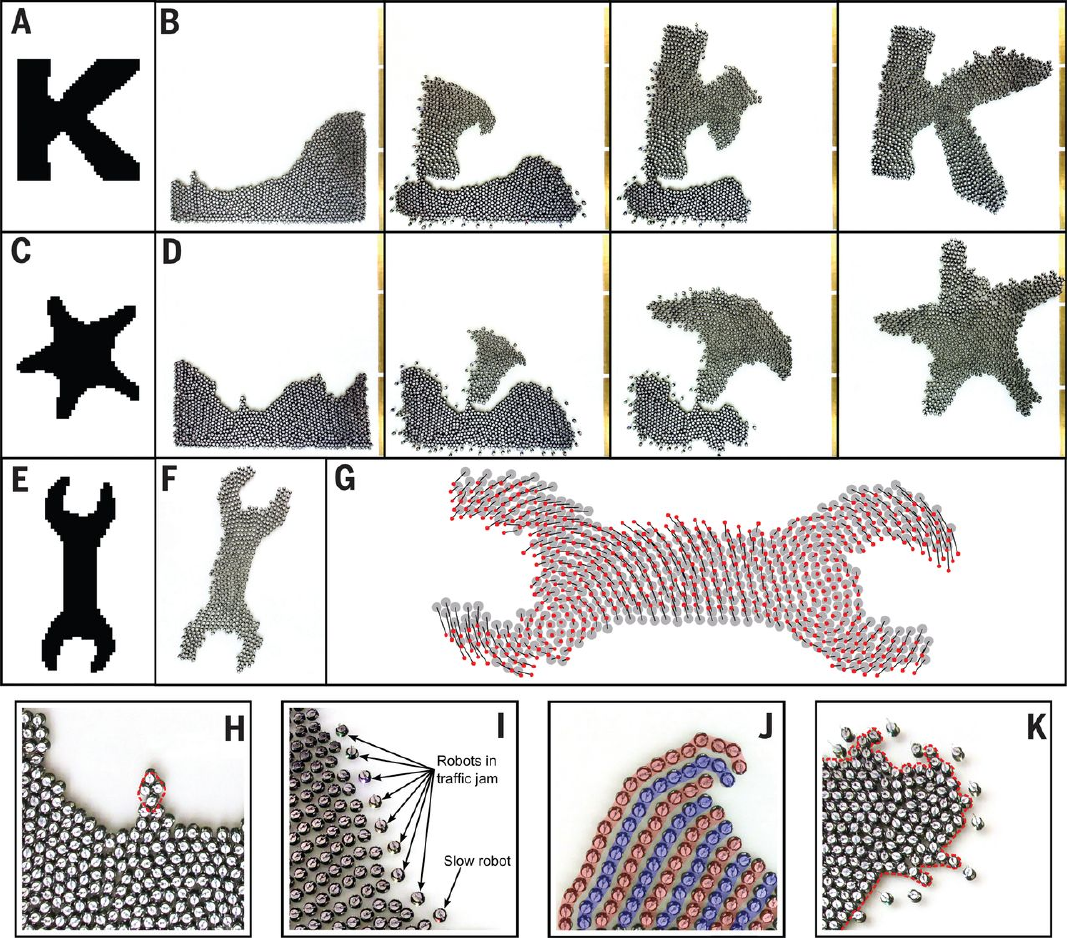}
    \caption{\label{fig:kilobots}Self-assembly experiments using up to 1024 physical robots~\cite{RubensteinEtAl2014}. (A, C, and E) Desired shape provided to robots as part of their program. (B and D) Self-assembly from initial starting positions of robots (left) to final self-assembled shape (right). Robots are capable of forming any simply connected shape, subject to a few constraints to allow edge-following (19). (F) Completed assembly showing global warping of the shape due to individual robot errors. (G) Accuracy of shape formation is measured by comparing the true positions of each robot (red) and each robot’s internal localized position (gray). (H to K) Close-up images of starting seed robots (H), traffic backup due to a slowly moving robot (I), banded patterns of robots with equal gradient values after joining the shape (robots in each highlighted row have the same gradient value) (J), and a complex boundary formed in the initial group (dashed red line) due to erosion caused by imprecise edge-following (K).}
\end{figure}

Robots are physically situated~\cite{Brooks:1991,steels1995artificial} and embedded~\cite{BeerInPress} in their environment, so they have been useful to explore aspects of life related to behavior~\cite{Maes1994}, traditionally studied by ethology~\cite{Beer1990}. Even when there are also several soft ALife models of adaptive behavior \cite{Braitenberg:1986,Gershenson2004}, dealing with physics (time, motion, inertia, gravity, hardware imperfections, etc.) proves to be already an important challenge for ALife. As mentioned earlier, robotic controllers have usually been evolved in software that has been uploaded into hardware. This is also known as ``evolutionary robotics''~\cite{Cliff1993,Nolfi2000,Harvey2005}.  Hard ALife models have also been used extensively to study the emergence of collective behaviors~\cite{DorigoEtAl2004,ZykovEtAl2005,Halloy2007,RubensteinEtAl2014,Vasarhelyi2018} (Figure~\ref{fig:kilobots}).

\section{Wet ALife}

Wet ALife is related to the effort of creating artificial cells in the laboratory from chemical and biochemical precursors. Living cells are the basic structural, functional, and biological units of all known living organisms. They are found in nature and produced and maintained by homeostasis, self-reproduction and evolution~\cite{luisi2002toward}. In contrast, artificial cells (synthetic cells) are prepared by humans and only mimic some of the properties, functionalities, or processes of natural cells. 

Although there are many laboratories working on this task, the successful preparation of artificial cells having all features of natural living cells is still challenging and an artificial cell that is able to self-produce and maintain itself (so-called autopoietic system) has not yet been demonstrated. All published papers with “artificial cell” in the title describe usually simple particles that have at least one property in common with living cells. Nevertheless, a big challenge exists to synthesize an artificial cell having at least several properties shared by living cells, \emph{e.g.} \emph{(i)} the presence of a stable semi-permeable \emph{membrane} that mediates the exchange of molecules, energy, and information between internal content and external environment while preserving specific identity, \emph{(ii)} the possibility to sustain themselves by using energy from its environment to manufacture at least some components from resources in the environment using \emph{metabolism} and \emph{(iii)} the capacity of growth and self-replication including genetic \emph{information}. 

A simpler and less problematic approach is to synthesize protocells that are not necessarily alive, but that exhibit only a few life-like properties~\cite{hanczyc2003experimental,Protocells2008,rasmussen2004transitions}. Protocells can be defined as simplified systems that mimic one or many of the morphological and functional characteristics of biological cells. Their structure and organization are usually very simple and can be orthogonal to any known living system. Protocells are used both to model artificial life and to model the origins of life. In the latter case, protocells are often considered as hypothetical precursors of the first natural living cells.

\begin{figure}[ht]
    \centering 
    \includegraphics[width=\textwidth]{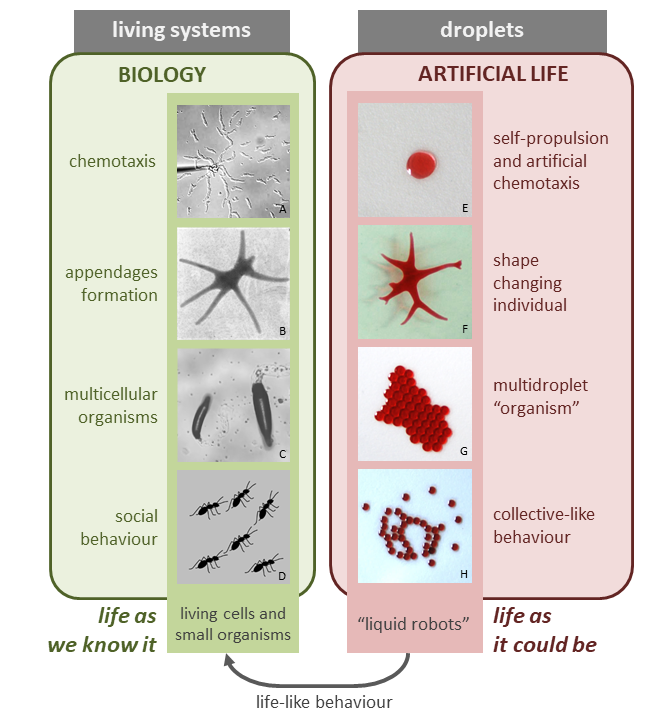}
    \caption{\label{fig:droplets}Schematic comparisson of wet artificial life research on droplets to biology and studies of living systems.   A – Chemotaxis of \emph{Dictyostelium} cells.  B - Prosthecate freshwater bacteria.  C – Multicellular slug stage of the \emph{Dictyostelium} developmental cycle.  D – Schematic representation of ants group.  E-H – Decanol droplets placed into aqueous solution of sodium decanoate.}
\end{figure}

In principle, two main motivations exist to create an artificial cell. One group of researchers aims to answer  questions about the origin of life: they synthesize primitive cells which consist of a protocell membrane that defines a spatially localized compartment, and of genetics polymers that allow for the replication and inheritance of functional information. The aim is to create self-replicating micelles or vesicles and to observe spontaneous Darwinian evolution of protocells in the laboratory~\cite{bachmann1992autocatalytic,walde1994autopoietic}. Other researchers want to prepare particles with life-like properties that mimic the behaviour of living cells, though without the ability to self-replicate or evolve. Such objects can move in their environment ~\cite{cejkova2014dynamics,hanczyc2007fatty}, selectively exchange molecules with their surrounding in response to a local change in temperature or concentration, chemically process those molecules and either accumulate or release the product, change their shape~\cite{vcejkova2018multi,cejkova2016evaporation} and behave collectively~\cite{cejkova2019dancing,Qiao:2017}. Such synthetic objects can be used for instance as smart drug delivery vehicles that release medicine \emph{in-situ}. These artificial cells could also be called chemical or liquid robots~\cite{vcejkova2017droplets} (Figure~\ref{fig:droplets}).

Recently, so-called Xenobots were introduced~\cite{Kriegman2020,Blackistoneabf1571}. Although they do not belong to traditional wet artificial life research, it should be mentioned here as a new approach on how to synthesise artificial organisms in the laboratory by using ALife tools. Xenobots were designed by an evolutionary computer algorithm and then assembled them from embryonic frog cells \emph{Xenopus laevis} (hence the name Xenobots). If Xenobots are “living robots”, “living machines” or "man-made animals" is still a debatable question. Nevertheless, these creatures, smaller than a millimeter in size, can find potential applications in areas such as environmental remediation. At this moment, Xenobots have the ability neither to self-replicate nor to evolve, but the life span and the hypothetical ability to reproduce could be assessed and regulated in the future in accordance with ethical principles.

\section{Self-replicating systems}

Let us imagine a factory where robots are constructing other robots. We know such scenes from the movie \emph{I, Robot} where the company U. S. Robotics produces humanoid robots, or from an island factory in the theatre play \emph{R.U.R. – Rossum’s Universal Robots}. However, these examples are fictional and such factories have never existed in the real world and still belong to science fiction. Still, the idea of machines that  beget machines was already in the air four centuries ago. A comprehensive study on the history and state of art of self-replicating machines from a scientific and technological point of view can be found in a recent book by Taylor and Dorin~\cite{taylor2020rise}.

Self-replicating systems can be categorized as follows. “Standard-replicators” that could reproduce by building a copy of themselves. Machines that are able not only to reproduce, but also evolve by natural selection as their living counterparts, are categorized as “evolvable self-replicators” (evo-replicators). And so-called “manufacturing self-replicators” (maker-replicators) have the ability not only to self-replicate, but also to create specific goods and materials as by-products when they self-replicate.

\section{Challenges and opportunities: a 2050 vision}

ALife will have to tackle very specific challenges in science and engineering: \emph{(a)} Get the agents out of the lab and into the wild; \emph{(b)} Make them able to interact with living organisms in a predictable way; \emph{(c)} Make them sustainable and degradable in order not to deteriorate the already transformed environments; \emph{(d)} Allow long-term operations of these agents, or even multi-generational time spans by self-reproduction of the agents. The development of sustainable technologies is thus urgently needed, and ALife could in part address that. The ALife systems should be characterized by robustness, autonomy, energy efficiency, materials recycling, local intelligence, self-repair, adaptation, self-replication, and evolution, all properties that traditional technologies lack, but living systems possess~\cite{Bedau:2009,Bedau2013Introduction-to}.

Concerning \textbf{soft ALife}, perhaps one of the greatest challenges is that of open-ended evolution (OEE)~\cite{standish2003open,MorenoRuiz2006,taylor2016open,e22101163}: can a program produce ever-increasing complexity (as it seems natural evolution does)? Hern\'andez-Orozco \emph{et al}.~\cite{hernandez2018undecidability} have proposed that undecidability and irreducibility are requirements for OEE. It might be that the difficulty of achieving OEE is related to the limits of formal systems~\cite{godel1931formal,Turing:1936,Chaitin1975}. Simplifying the situation: formal systems cannot change their own axioms. This is a necessary condition for traditional logic, mathematics, and computation, but perhaps OEE requires precisely the possibility of changing axioms.

As for \textbf{hard ALife}, robots are becoming more and more sophisticated~\cite{Rahwan2019}. They are also gaining in autonomy. However, as with the rest of artificial intelligence, all robots are specialists. They are good to perform the tasks they were designed for, but they cannot generalize and perform other activities. For example, a vacuum robot cannot paint. Not only because it lacks the appropriate hardware, but also the software is task-specific. The so-called ``artificial general intelligence'' has so far produced not more than mere speculations. Could it be that the limits of formal systems just mentioned for OEE also affect artificial intelligence in general and robots in particular? If so, can we find an alternative, to build robots that are not based on formal systems?

Perhaps the most promising is the least developed: \texttt{wet ALife}. If we build a system that most people agree on calling ``alive'', most probably it will be from wet ALife. There are several challenges already mentioned, but there seems to be no inherent limit for building or finding alternative life forms, either artificial or extraterrestrial. It is a blind guess to try to say whether we will have detected or created life different from the one that evolved on Earth by 2050. Still, soft ALife and hard ALife seem to have inherent limits (derived from the limits of formal systems), so we might as well expect the most from wet ALife. In any case, it can certainly contribute to a ``general biology''.

Physics can also contribute in this direction. Already, research in self-organization~\cite{NicolisPrigogine1977,MartyushevSeleznev2006} and active matter~\cite{ramaswamy2017active} have contributed to understanding properties of living systems. Very probably, in the next few decades, advances in physics will enhance our perspectives on what we consider to be living, how it evolved, and where it might lead to.

There is still no agreed definition of life. Biological systems are perhaps too complex for a sharp definition. As we have seen, ALife can help to understand general properties of living systems. This can benefit biology and engineering, gaining insights on life and being able to build artificial systems exhibiting properties of the living.

\bibliographystyle{unsrt}
\bibliography{References_01_03_03}

\end{document}